\documentclass[runningheads]{llncs}
\usepackage{graphicx}
\usepackage{booktabs}
\usepackage{algorithm,algpseudocode}
\usepackage{bbding}
\usepackage{hyperref}
\usepackage{url}
\usepackage{booktabs}
\usepackage{amsfonts}
\usepackage{nicefrac}
\usepackage{microtype}
\usepackage{xcolor}
\usepackage{xspace}
\usepackage{multirow}
\usepackage{multicol}
\usepackage{pifont}
\usepackage{xparse}

\usepackage{caption}
\usepackage{subcaption}

\newcommand{\cmark}{\ding{51}}%

\usepackage{tikz}
\usepackage{comment}
\usepackage{amsmath,amssymb} 
\usepackage{color}

\usepackage[accsupp]{axessibility}  

\newcommand\blfootnote[1]{%
	\begingroup
	\renewcommand\thefootnote{}\footnote{#1}%
	\addtocounter{footnote}{-1}%
	\endgroup
}

\begin{document}
\pagestyle{headings}
\mainmatter
\def\ECCVSubNumber{1069}  
\vspace{-1cm}
\title{Entropy-driven Sampling and Training Scheme \\ for Conditional Diffusion Generation}
\vspace{-0.45cm}
\titlerunning{ED Sampling and Training Scheme for Conditional Diffusion Generation}


\newcommand*\samethanks[1][\value{footnote}]{\footnotemark[#1]}

\author{
Shengming Li\inst{1}\thanks{The first two authors contributed equally to this paper.} \and
Guangcong Zheng\inst{1}\samethanks[1] \and
Hui Wang\inst{1} \and 
Taiping Yao\inst{2} \and
Yang Chen\inst{2} \and
Shouhong Ding\inst{2} \and
Xi Li\inst{1,3,4}\thanks{The corresponding author is Xi Li.}
}

\authorrunning{S. Li \& G. Zheng et al.}

%
\institute{
College of Computer Science \& Technology, Zhejiang University \\
    \email{\{guangcongzheng, shengming22, wanghui\_17, xilizju\}@zju.edu.cn}
\and Youtu Lab, Tencent, China \\
    \email{\{taipingyao, wizyangchen, ericshding\}@tencent.com}
\and Shanghai Institute for Advanced Study, Zhejiang University 
\and Shanghai AI Laboratory
}
\maketitle

\newcommand{\grad}{\nabla}
\newcommand{\E}{\mathbb{E}}
\newcommand{\Var}{\mathrm{Var}}
\newcommand{\Cov}{\mathrm{Cov}}
\newcommand{\Ea}[1]{\E\left[#1\right]}
\newcommand{\Eb}[2]{\E_{#1}\!\left[#2\right]}
\newcommand{\Vara}[1]{\Var\left[#1\right]}
\newcommand{\Varb}[2]{\Var_{#1}\left[#2\right]}
\newcommand{\kl}[2]{D_{\mathrm{KL}}\!\left(#1 ~ \| ~ #2\right)}
\newcommand{\pdata}{{p_\mathrm{data}}}
\newcommand{\bA}{\mathbf{A}}
\newcommand{\bI}{\mathbf{I}}
\newcommand{\bJ}{\mathbf{J}}
\newcommand{\bH}{\mathbf{H}}
\newcommand{\bL}{\mathbf{L}}
\newcommand{\bM}{\mathbf{M}}
\newcommand{\bQ}{\mathbf{Q}}
\newcommand{\bR}{\mathbf{R}}
\newcommand{\bzero}{\mathbf{0}}
\newcommand{\bone}{\mathbf{1}}
\newcommand{\bb}{\mathbf{b}}
\newcommand{\bg}{\mathbf{g}}
\newcommand{\bu}{\mathbf{u}}
\newcommand{\bv}{\mathbf{v}}
\newcommand{\bw}{\mathbf{w}}
\newcommand{\bx}{\mathbf{x}}
\newcommand{\by}{\mathbf{y}}
\newcommand{\bz}{\mathbf{z}}
\newcommand{\bxh}{\hat{\mathbf{x}}}
\newcommand{\btheta}{{\boldsymbol{\theta}}}
\newcommand{\bphi}{{\boldsymbol{\phi}}}
\newcommand{\bepsilon}{{\boldsymbol{\epsilon}}}
\newcommand{\bgamma}{{\boldsymbol{\gamma}}}

\newcommand{\ie}{{\emph{i.e.}},\xspace}
\newcommand{\viz}{{\emph{viz.}},\xspace}
\newcommand{\eg}{{\emph{e.g.}},\xspace}
\newcommand{\etc}{etc.}
\newcommand{\etal}{{\emph{et al.}}}
\newcommand{\wrt}{{\emph{w.r.t.}}}

\newcommand{\hbepsilon}{{\boldsymbol{\hat \epsilon}}}
\newcommand{\bmu}{{\boldsymbol{\mu}}}
\newcommand{\bnu}{{\boldsymbol{\nu}}}
\newcommand{\bSigma}{{\boldsymbol{\Sigma}}}

\begin{abstract}
    Denoising Diffusion Probabilistic Model (DDPM) is able to make flexible conditional image generation from prior noise to real data, by introducing an independent noise-aware classifier to provide conditional gradient guidance at each time step of denoising process.
    However, due to the ability of classifier to easily discriminate an incompletely generated image only with high-level structure, the gradient, which is a kind of class information guidance, tends to vanish early, leading to the collapse from conditional generation process into the unconditional process.
    To address this problem, we propose two simple but effective approaches from two perspectives.
    For sampling procedure, we introduce the entropy of predicted distribution as the measure of guidance vanishing level and propose an entropy-aware scaling method to adaptively recover the conditional semantic guidance. 
    For training stage, we propose the entropy-aware optimization objectives to alleviate the overconfident prediction for noisy data.
    On ImageNet1000 256$\times$256, with our proposed sampling scheme and trained classifier, the pretrained conditional and unconditional DDPM model can achieve 10.89\% (4.59 to 4.09) and 43.5\% (12 to 6.78) FID improvement respectively. 
    Code is available at \url{https://github.com/ZGCTroy/ED-DPM}.
\end{abstract}
\section{Introduction}
Conditional image generation, usually class conditional, aims to generate the specific class of high-quality images.
There are a lot of generative models which are able to make high-quality conditional generation based on joint-training scheme, like Generative Adversial Networks (GAN) \cite{bigbigan,biggan} or Variational Autoencoder (VAE) \cite{vqvae2,nvae}.
However, when the condition requirement is changed, the generative models will be retrained, which is very inconvenient.
\blfootnote{
$^{*}$Equal contribution \\
Preprint. Under review.}

\begin{figure}[t!]
    \begin{center}
    \includegraphics[width=0.87\textwidth]{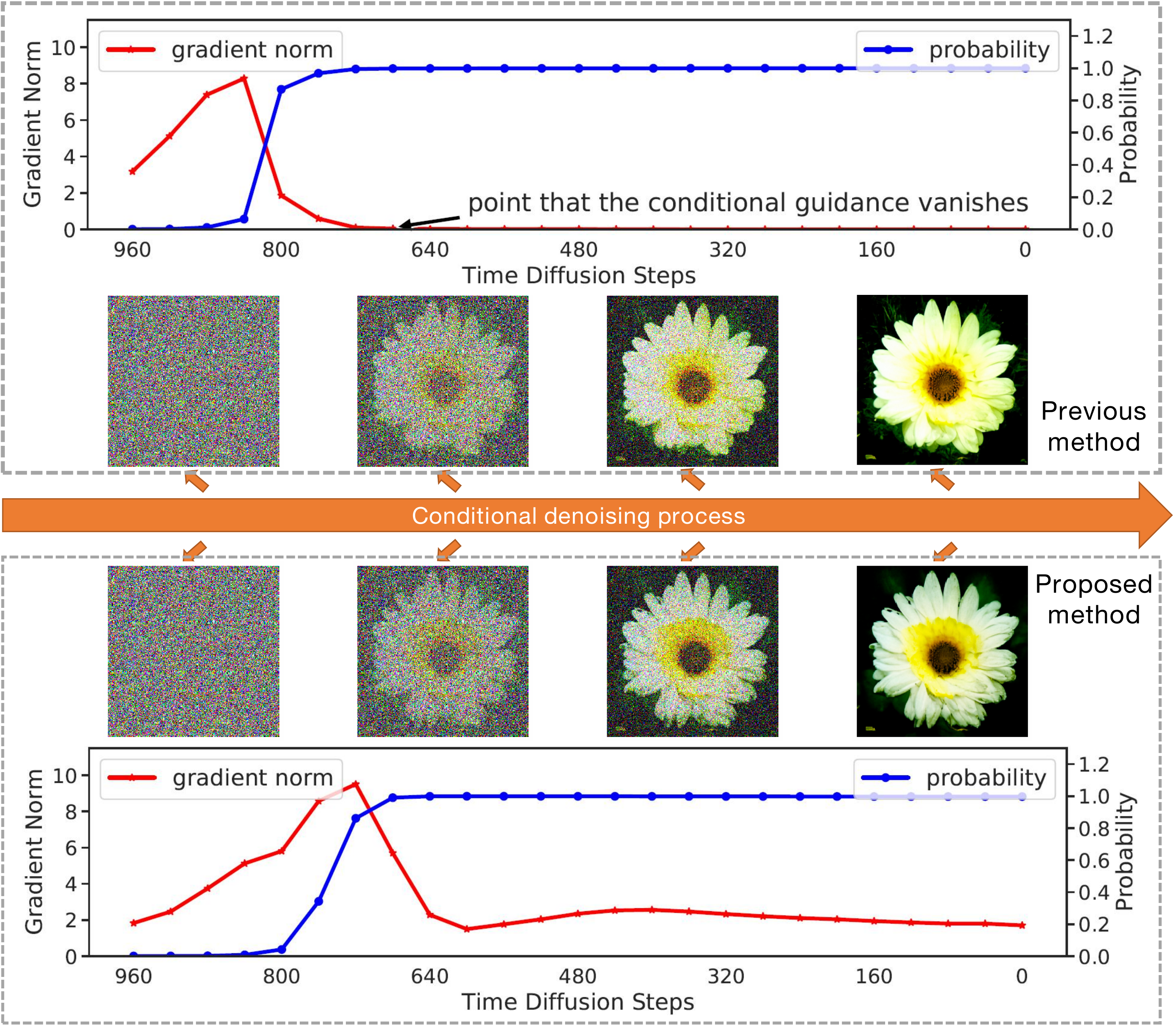}
    \caption{
    The visualization of denoising sampling process. The classifier gradient, a kind of class information in conditional generation, quickly converge to 0 in the previous method. It will lead to the collapse from conditional generation to unconditional generation, while our method recovers the gradient guidance and succeed to generate fine-grained features in the subsequent iterations.
    }\label{fig:coverge}
   
    \end{center}
    \vskip -0.2in
\end{figure}

Denoising Diffusion Probabilistic Model (DDPM) \cite{ho2020denoising,nichol2021improved,song2020denoising} is a class of iterative generation models, which has made remarkable performance in unconditional image generation recently.
The flexibility of DDPM  \cite{dhariwal2021diffusion,dickstein} is that it can be easily extended to conditional variants by introducing an independent noise-aware classifier.
Recent researches modeled the prior denoising distribution by training an unconditional DDPM, following the training scheme of Denoising Score Matching \cite{vincent2011connection}, and computed likelihood score by backwarding the classifier gradient.
Dhariwal \etal~\cite{dhariwal2021diffusion} further proposed fixed scaling factor to improve the predicted probability of generated samples for DDPM, achieving superior performance than GAN on several image generation benchmarks.

In conditional generation process of DDPM, by backwarding the gradient of classification probability to image, the classifier provides high-level semantic information in the early stage of iterations, and gradually strengthens fine-grained features in the subsequent iterations, both of which are indispensable. 
However, there exists a huge gap between discriminating the class of a image and generating a specific class of image with fine-grained textures. As shown in Fig.\ref{fig:coverge}, the predicted distribution of the classifier for noisy images tends to quickly converge to the desired class distribution, which is one-hot distribution, leading to the early vanishing of conditional gradient guidance. This is because that the incompletely generated image, which is still a noisy image and lacks fine-grained features, can be easily classified in the middle of denoising process.
In this way, the image is considered to have been completely generated, and will no longer be guided by classifier gradient containing class information. As a result, the conditional generation process will degrade into an unconditional generation process in the later stage.

Therefore, our motivation is to enable the classifier to continuously give conditional guidance throughout the entire denoising process. We propose two simple but effective schemes from the procedure of sampling and the design of classifier training.

From the perspective of sampling procedure, we focus on how to detect the gradient vanishing and rescale the gradient to avoid the existence of gradient vanishing or recover the gradient when the vanishing does happen. 
We propose \textbf{E}ntropy-\textbf{D}riven conditional \textbf{S}ampling (EDS) method, which is able to adaptively measure the level of gradient vanishing and rescale the gradient guidance to a appropriate level. 
In design of training classifier, we propose \textbf{E}ntropy-\textbf{C}onstraint \textbf{T}raining (ECT), which will penalize the classifier when it gives a overconfident classification probability to a generated noisy image, thus constraining the classifier to provide more gentle guidance.

Our contributions can be summarized as follows:
\begin{itemize}
    \item[$\bullet$] We are the first to discover the problem of vanishing gradient guidance for ddpm-based conditional generation methods, and point out that category information guidance should be continuously provided throughout the entire generation process.
    \item[$\bullet$]  We propose EDS to alleviate the vanishing guidance by dynamically measuring and rescaling gradient guidance.
    At the training stage of classifier, for alleviating the vanishing gradient caused by one-hot label supervision, we utilize discrete uniform distribution to build an entropy-aware optimization term, which is \textbf{E}ntropy-\textbf{C}onstraint \textbf{T}raining scheme (ECT).
	\item[$\bullet$] We conduct experiments on ImageNet1000 and achieve state-of-the-art FID (Fréchet Inception Distance) results at various resolutions. On ImageNet1000 256$\times$256, with our proposed sampling scheme and trained classifier, the pretrained conditional and unconditional DDPM model can achieve 10.89\% (4.59 to 4.09) and 43.5\% (12 to 6.78) FID (Fréchet Inception Distance) improvement respectively. 
\end{itemize}

\section{Related Work}
\subsection{Denoising Diffusion Probabilistic Model}
Denoising diffusion probabilistic models (DDPM) is the latest generation model which achieve superior generation performance than traditional generative models like Generative Adversial Networks (GAN) \cite{gan,bigbigan,biggan}, Variational Autoencoders (VAE) \cite{nvae,vqvae,vqvae2} on several benchmarks about unconditional generation.
Its key idea is to model the diffusion process based on total $T$ time steps, which adds noise gradually to the clean data, and its reverse process, which denoises the white noise into the clean sample.
Accordingly, its diffusion process is modeled as a fixed Markov Chain and its transition kernel is formulated as:
\begin{equation}\label{eq:rw_forward}
q(\bx_{t}|\bx_{t-1}):=\mathcal{N}(\bx_{t};\sqrt{1-\beta _{t}}\bx_{t-1},\beta _{t}\mathbf{I}),
\end{equation} 
where $\beta_1,...,\beta_T$ are the fixed variance parameters, which are not learnable.
According to the transition kernel above, when the clean data $\bx_0$ is given, the noisy data $\bx_t$ can be sampled with a close-formed distribution:
\begin{equation}\label{eq:sample_diffusion}
q(\bx_{t}|\bx_{0})=\mathcal{N}(\bx_{t};\sqrt{\overline{\alpha}_{t}}\bx_{0},(1-\overline{\alpha}_{t})\mathbf{I}),
\end{equation}
where $\alpha_t=1-\beta_t$ and $\bar \alpha_t=\prod_{s=1}^t \alpha_s$.
When $t$ is close to $T$, $x_T$ can be approximated as a Gaussian distribution.

Given a prior diffusion process, DDPM aims to model its reverse process to sample from the data distribution.
The optimization objective of the reverse transition can be derived from a variational bound \cite{kingma2014auto}.
Thus, DDPM introduced the variational solution \cite{kingma2014auto} and assumed that its reverse transition kernel also subjects to Gaussian distribution, which is the same as the diffusion process.
In this way, the generation process parameterized the mean of the Gaussian transition distribution and fixed its variance as follow:
\begin{equation}\label{eq:reverse}
\begin{split}
p_{\theta}(\bx_{t-1}|\bx_{t})=\mathcal{N}(\bx_{t-1};\mu_{\theta}(\bx_t),{\sigma}_t^2\mathbf{I}) 
\\
\mu_{\theta}(\bx_t)=\frac{1}{\sqrt{\alpha_{t}}}(\bx_{t}-\frac{1-\alpha _{t}}{\sqrt{1-\bar {\alpha}_{t}}}\bepsilon_{\theta }(\bx_{t})),
\end{split}
\end{equation}
where $\bepsilon_{\theta }(\bx_{t})$ is a noise estimator modeled by a neural network.
The variance is designed as the hyperparameters for training diffusion models.

Recently, there were some researches \cite{scorematching,sde,ddim,dhariwal2021diffusion} which indicated that the noise estimator can be regarded as an approximation of score function so that the sampling process is equivalent to solving a stochastic differential equation.
Based on above, Song.et \cite{ddim} proposed an effective sampling process that shares the same training objectives as DDPM and its corresponding denoising process, which is also the iteration solution to solve the stochastic differential equation, is designed as followed:
\begin{equation}\label{eq:ddim_sample}
\bx_{t-1}=\sqrt{\alpha_{t-1}}f_{\theta}(\bx_t, t) + \sqrt{1-\alpha_{t-1}-\sigma_t^2} \epsilon_{\theta}(\bx_t) + \sigma_t^2 \mathbf{\bz},
\end{equation}
where $f_{\theta}(\bx_t,t)$ is the prediction of clean data $x_0$ when the noisy data $\bx_t$ is observed and noise prediction $\bepsilon_{\theta}(\bx_t)$ is given.
The specific form of $f_{\theta}(\bx_t,t)$ can be expressed as:
\begin{equation}\label{eq:ddim_predx0}
f_{\theta}(\bx_t, t)=\frac{\bx_t-\sqrt{1-\alpha_t}\bepsilon_{\theta}(\bx_t)}{\sqrt{\alpha_t}},
\end{equation}
When the variance $\sigma_t$ is set to 0, the sampling process becomes deterministic.
At the same time, the non-Markovian diffusion process \cite{ddim} allows that the generation quality remains unchanged within fewer denoising steps, which is called DDIM sampling process.

\subsection{Conditional Image Generation}
Conditional image generation aims to generate samples with desired condition information.
The condition can be extended to multi-modal information, such as class \cite{cgan,vqvae,vqvae2,xiao2021conditional}, text \cite{xia2021tedigan,ramesh2021zero,wang2021cycleconsistent}, and low-resolution image \cite{bulat2018learn}.
Most previous work modeled this by the joint training scheme with both condition and random noise, utilizing generative models like GAN or VAE.

Considering that Denoising Diffusion Probabilistic Model (DDPM) has made remarkable progress in recent years \cite{ddim,kingma2021variational,ddimdistill,huang2021variational,de2021diffusion,song2021maximum,nachmani2021non,nie2021controllable}, there raised many researches \cite{dhariwal2021diffusion,diffwave,choi2021ilvr,sasaki2021unit,sinha2021d2c,liu2021diffsinger,lyu2021conditional} which applied DDPM to conditional generation.
Due to the ideal theoretical properties of DDPM, it can be extended flexibly to conditional variants utilizing Bayes theorem without retraining, which is similar with score-based generative models \cite{scorematching,sde,meng2021sdedit}.
In this paper, we focus on the class-conditional generation task,
in which the condition is represented by a class discrete distribution, and design a more effective sampling and training scheme to improve the generation quality for DDPM, further exploring its potential in image generation aspects.


\section{Proposed Methods}
We start by introducing the conditional generation process for diffusion models with classifier guidance (Sect.(\ref{md:cdg})).
For ease of description, we will firstly introduce our dynamic scaling technology to recover gradient guidance adaptively in the sampling process and its motivation (Sect.(\ref{md:eds})).
Then, we describe the entropy-aware optimization loss for alleviating vanishing conditional guidance from training perspective (Sect.(\ref{md:ect})), utilizing the uniform distribution, which is a more dense distribution.

\subsection{Conditional Diffusion Generation}
\label{md:cdg}



\begin{figure}[t]
    \begin{center}
    \includegraphics[width=1.\textwidth]{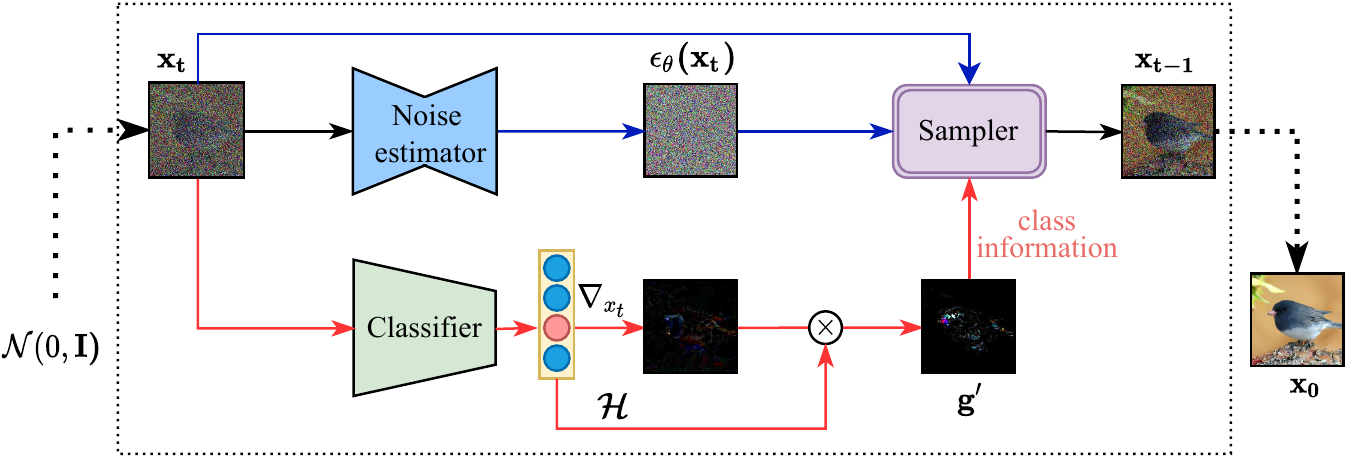}
    \caption{Pipeline for Entropy-driven Sampling process. Sampler represents a class of iteration method (DDPM or DDIM), which is non-parametric. All models are pretrained without gradient updating in the sampling process.}\label{fig:pipeline}
    \end{center}
    \vskip -0.2in
\end{figure}

The goal of conditional image generation is to model probability density $p(\bx|\by)$.
To be specific, in class-conditional image generation, $\by$ represents the desired class label, which tends to be one-hot distribution, and $\bx$ represents the image sample.
For diffusion models, it can be implemented by introducing an independent classifier, as shown in Fig.\ref{fig:pipeline}.

Specifically, the goal is converted into modeling the conditional transition distribution $p(\bx_{t-1}|\bx_{t},\by)$, derived from Markov chain sampling scheme:
\begin{equation}\label{eq:cond_markov}
\begin{split}
p_{\varphi}(\bx_{0}|\by)&=\int_{}^{}p_{\varphi}(\bx_{0:T}|\by)d\bx_{1:T}, \\
p_{\varphi}(\bx_{0:T}|\by)&=p(\bx_{T})\prod_{t=1}^T p_{\varphi}(\bx_{t-1}|\bx_{t},\by),
\end{split}
\end{equation}
where $\varphi$ represents the model and $T$ is the total length of Markov chain, which tends to be large.
Then, we decompose the conditional transition distribution into two independent terms:
\begin{align}\label{eq:cond_kernel_decompose}
\begin{split}
p_{\varphi}(\bx_{t-1}|\bx_t,\by)=Zp_{\theta}(\bx_{t-1}|\bx_{t})p_{\phi}(\by|\bx_t),
\end{split}
\end{align}
where $Z$ is a normalizing constant independent from $\bx_{t-1}$ and $\varphi$ can be seen as the combination of models $\theta$ and $\phi$,
which is proven theoretically \cite{dickstein,dhariwal2021diffusion}.
Furthermore, the log density of Eq.\eqref{eq:cond_kernel_decompose} can be approximated as a Gaussian distribution \cite{dhariwal2021diffusion}:
\begin{equation}\label{eq:cond_gaussian_format}
\begin{split}
&\log (p_{\varphi}(\bx_{t-1}|\bx_t,\by)) \approx \log p(\bz) + \log Z \\
\bz &\sim \mathcal{N}(\frac{1}{\sqrt{\alpha_{t}}}(\bx_{t}-\frac{1-\alpha _{t}}{\sqrt{1-\bar {\alpha}_{t}}}\bepsilon_{\theta }(\bx_{t})) + \sigma^2_t \bg,{\sigma^2_t} \mathbf{I}),
\end{split}
\end{equation}
where $\bg=s\nabla_{\bx_t} \log p_{\phi}(\by|\bx_t)$ and $s$ is the gradient scale.
$\bg$ can be derived by backwarding the gradient of a pretrained classifier about noisy data $\bx_t$.
Usually, $s$ is set to a constant \cite{dhariwal2021diffusion} to improve the predicted probability.
$\bepsilon_{\theta }(\bx_{t})$ is the pretrained noise estimator for diffusion models and 
$\sigma^2_t$ is the hyperparameters for unconditional DDPM to control the variance.

In this way, unconditional diffusion models with parameterized noise estimator $\bepsilon_{\theta}$ can be extended to conditional generative models by introducing the condition-aware guidance $\nabla_{\bx_t} \log p_{\phi}(\by|\bx_t)$, 
as shown in Fig.\ref{fig:pipeline}.
Specifically, we start with the prior noise $\bx_T \sim \mathcal{N}(0,\bI)$, and utilize DDPM-based (Eq.\eqref{eq:cond_gaussian_format}) sampler to make transition iteratively to generate the samples conditioned on desired class $\by$.
The sampler can also be extended into DDIM iteration method (Eq.\eqref{eq:reverse}). The extension details for conditional process with classifier guidance can refer to Dhariwal \etal \cite{sde,dhariwal2021diffusion}.


\subsection{Entropy-driven Conditional Sampling}
\label{md:eds}

\begin{figure}[t!]
    \begin{center}
    \includegraphics[width=1.0\textwidth]{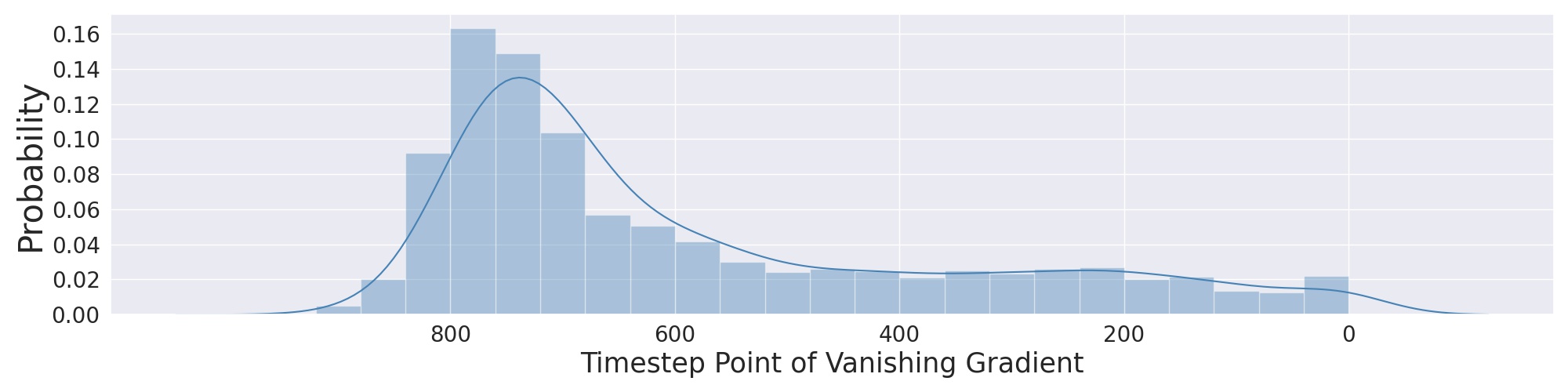}
    \caption{The various gradient vanishing points for different samples. We randomly generate 5000 samples and define the time point, in which the gradient norm is smaller than 0.15, as the gradient vanishing point. Premature vanishing of gradients occurs in almost every sample.}\label{fig:vanishing_hist}
    \end{center}
    \vskip -0.3in
\end{figure}
During the conditional generation process, we observe that the guidance provided from noise-aware classifier tends to vanish prematurely,
which could be attributed to the discrepancy between generative pattern and discriminative pattern.
For example, the noisy samples with high-level semantic information, such as contour or the color, may be guided iteratively by the classifier with nearly one-hot distribution,
in which situation the gradient guidance tends to be weak or vanish, while the samples still lack the condition-aware semantic details. 
In this way, the condition-aware textures of generated samples are guided by unconditional denoising process (Eq.\eqref{eq:reverse}).

An intuitive solution is to manually select a time step during the sampling process, after which the semantic details tend to vanish in most instances, and rescale the conditional gradient by an empirical constant after the selected time step.
However, since the stochasticity in the generation process of diffusion models \cite{choi2021ilvr},
the denoising trajectories for generated samples would differ from each other. 
It will lead to the various initial vanishing points, as shown in Fig.\ref{fig:vanishing_hist}.
At the same time, considering the learning bias of classifier for different conditional classes, the level of recovery factor for each class may also differ.
In summary, the effective scaling factor can be related to the current time step, the class condition, and the stochasticity of generation process.
The experiment design and results about more intuitive approaches can be seen in Sect.\ref{exp:eds_ab}.

Motivated from above, we propose an dynamic scaling technology for the conditional diffusion generation to recover the semantic details adaptively for each sample.
We noticed that, when time step is close to $T$, the predicted distribution tends to be dense, which can be nearly approximated as uniform distribution.
The reason is that the noisy data derived from Eq.\eqref{eq:rw_forward} can be approximated as random noise, $\mathcal{N}(0,\bI)$,
in which state the gradient guidance is obvious.
As time step declines to $0$ inversely, the noise hidden in sample will be removed gradually and the predicted distribution tends to be close to one-hot, in which case the classifier gradient is invalid to provide semantic details for generation.
Statistically, entropy can represent the sparsity of the predicted distribution, inspiring us to take it into consideration:
\begin{equation}\label{eq:condentropy}
\begin{split}
\mathcal{H}(p_{\phi}(\tilde \by|\bx_t)) &= - \E_{\tilde \by|\bx_t} \log p_{\phi}(\tilde \by|x_t) \\
 &= - \sum_{i=1}^{|Y|} p_{\phi}(\tilde \by_i|\bx_t) \log p_{\phi}(\tilde \by_i|\bx_t),
\end{split}
\end{equation}
where $p_{\phi}(\tilde \by|\bx_t)$ represents the predicted distribution from classifier and $Y$ represents the set of all class conditions.

  
\begin{algorithm}[t]
    \caption{ \small{Entropy-driven sampling scheme (DDPM/DDIM)}}
    \label{alg:entropy_driven_ddpm}
        
    \begin{algorithmic}[1]
        \Require a pretrained diffusion model $\epsilon_{\theta}(\bx_t)$, classifier $p_{\phi}(\hat \by|\bx_t)$, and desired class condition $\by$
        \State $\bx_T \sim \mathcal{N}(\bzero, \bI)$
        \For{$t = T, \dotsc, 1$}
            \State $s \gets \gamma * \frac{\mathcal{H}(\mathcal{U}(\hat \by))}{\mathcal{H}(p_{\phi}(\hat \by|\bx_t))}$ if EDS, else $s \gets \gamma$
    
            \If{use DDPM}
                \State $\bz \sim \mathcal{N}(\bzero, \bI)$ if $t > 1$, else $\bz \gets \bzero$
                
                \State $\bg \gets  s \cdot \nabla_{\bx_t} \log p_{\phi}(\by|\bx_t),$
                
                \State $\bx_{t-1} \gets \frac{1}{\sqrt{\alpha_t}}\left( \bx_t - \frac{1-\alpha_t}{\sqrt{1-\bar\alpha_t}} \bepsilon_\theta(\bx_t) \right) + \sigma^2_t \bg + \sigma_t \bz$
            \ElsIf{use DDIM}
                \State $\hbepsilon \gets \bepsilon_{\theta}(\bx_t) -  s \cdot \nabla_{\bx_t} \log p_{\phi}(\by|\bx_t),$
                
                \State $\bx_{t-1} \gets \sqrt{\bar \alpha_{t-1}}(\frac{\bx_t - \sqrt{1- \bar \alpha_t}\hbepsilon}{\sqrt{\bar \alpha_t}}) 
                + \sqrt{1-\bar \alpha_{t-1}}\hbepsilon$
            \EndIf
        \EndFor
        \State \textbf{return} $\bx_0$
        
    \end{algorithmic}

\end{algorithm}

    
            


In this paper, we utilize $\mathcal{H}(p_{\phi}(\tilde \by|\bx_t))$ to adaptively fit various gradient vanishing time step.
Furthermore, the entropy $\mathcal{H}(p_{\phi}(\tilde \by|\bx_t))$ can also capture bias caused by different conditions, due to its sample-aware rescaling effect.
Thus, when we sample from a pretrained noise estimator $\bepsilon_{\theta}$ in Eq.(\eqref{eq:cond_gaussian_format}) conditionally, 
we reformulate the gradient term $\bg$ as following, which is shown in Fig.\ref{fig:pipeline}:
\begin{equation}\label{eq:condsample}
\begin{split}
\bg' &= s(x_t, \phi) * \nabla_{\bx_t} \log p_{\phi}(\by|\bx_t), \\
s(x_t, \phi) &= \gamma * \frac{\mathcal{H}(\mathcal{U}(\tilde \by))}{\mathcal{H}(p_{\phi}(\tilde \by|\bx_t))}
\end{split}
\end{equation}
where 
$\bgamma$ is a hyper-parameter to balance the guiding gradient and entropy-aware scaling factor $s(\bx_t,\phi)$.
In order to maintain the numerical range, we renormalize the entropy by its theoretical upper bound $\mathcal{H}(\mathcal{U}(\tilde \by))$, where $\mathcal{U}(\tilde \by)$ represents the uniform distribution of class variable, so that the gradients are almost not rescaled when $t$ is close to $T$.



\subsection{Training Noise-aware Classifier with Entropy Constraint}
\label{md:ect}
From the training perspective, the vanishing gradient can be partly attributed to the label supervision pattern for noise-aware classifier.
Since one-hot distribution is very sparse and is utilized to supervise the noisy data, the predicted distributions are inclined to converge to one-hot under noisy samples in the sampling process, so that the gradient guidances are too weak to generate condition-aware semantic details at sampling stage.
Specifically, given the dataset $(\bx_0, \by) \sim \mathbb{D}$ and a prior diffusion process (Eq.\eqref{eq:rw_forward}), the classifier will be trained under noisy data ${\bx_t}$ to build the gradient field in Eq.\eqref{eq:cond_gaussian_format} of each time step.
To alleviate the weak guidance caused by the sparsity, we utilize discrete uniform distribution, which is a dense distribution and has maximum entropy, as a perturbing distribution and introduce the optimization term at the training stage of the classifier to constrain the predicted distribution from the classifier as followed:
\begin{align}\label{eq:noisy_y}
\begin{split}
\mathcal{L}_{ECT}(\bx_t, \by) &= D_{KL}(p_{\phi}(\tilde \by|\bx_t) || \mathcal{U}(\tilde \by)) \\
&= \E_{\tilde \by|\bx_t} \log p_{\phi}(\tilde \by|\bx_t) - \E_{\tilde \by|\bx_t} \log \mathcal{U}(\tilde \by) 
\\
&= -\mathcal{H}(p_{\phi}(\tilde \by|\bx_t)) + \mathbf{C},
\end{split}
\end{align}
where $\mathbf{C}$ is a constant term independent from the parameter $\phi$.
This loss term is equivalent to maximizing entropy of the predicted distribution $p(\tilde \by|\bx_t)$.
The whole training loss of guiding classifier is composed of the normal cross-entropy loss and entropy constraint training loss (ECT), which is formally given by:
\begin{equation}\label{eq:noisy_training_loss}
\begin{split}
\mathcal{L}_{tot}(\bx_t,\by)=\mathcal{L}_{CE}(\bx_t,\by) + \eta  \mathcal{L}_{ECT}(\bx_t,\by),
\end{split}
\end{equation}
where $\eta$ is a hyper-parameter to adjust the divergence about
predicted label distribution and the uniform distribution.
\begin{algorithm}[!t]
    \caption{\small{Entropy-constraint training scheme.
    }}
    \label{alg:entropy_constraint_train}
    
    \begin{algorithmic}[1]
        \Require training set $\mathbb{D}$, a neural classifier $\phi$, training set $\mathbb{D}$, total time steps $T$
        \Repeat
            \State $(\bx_0,\by) \gets \text{sample from } \mathbb{D}$
            \State $t \sim \mathcal{U}(\{1, \dotsc ,T\})$
            \State $\bx_t \sim q(\bx_t|\bx_0)$ (Eq.\ref{eq:sample_diffusion})
            \State $L_{CE} \gets \by \log p_{\phi}(\tilde \by|\bx_t)$
            
            \If{use ECT}
                \State $L_{ECT} \gets - \mathcal{H}(p_{\phi}(\tilde \by|\bx_t)) $
                \State Take gradient descent step on $\nabla_{\phi} ( L_{CE} + \eta L_{ECT})$
            \Else
                \State Take gradient descent step on $\nabla_{\phi}  L_{CE}$
            \EndIf
        \Until{converged}
    \end{algorithmic}
\end{algorithm}

Different from Entropy-driven Sampling, the proposed training scheme tries to alleviate the vanishing guidance by adjusting the gradient direction in sampling process, instead of the gradient scale.
Thus, entropy-constrain training scheme can be complementary with entropy-driven sampling.

\section{Experiments}
In this section, we present experiments to verify the effectiveness and motivation of our proposed schemes. More visualization of generated samples and ablation experiments about hyperparameters can be found in supplementary materials.

\subsection{Experiment Setup}
\paragraph{\textbf{Dataset}}
We perform our experiments mainly on ImageNet dataset \cite{imagenet} at 256$\times$256 resolutions.
ImageNet contains 14,197,122 images with 1000 classes in total, which is a very challenging benchmark for conditional image generation.

\vspace{-2pt}
\paragraph{\textbf{Implementation Details.}}
For verifying the effect of proposed schemes, we apply the neural network architecture, ablated diffusion model (ADM), proposed by Dhariwal.et al \cite{dhariwal2021diffusion}. 
ADM is mainly based on the UNet, with increased depth versus width, the number of attention heads, and rescaling residual connections with $\frac{1}{\sqrt{2}}$.
It is also possible to train a conditional diffusion models. We call the conditional diffusion architecture \cite{dhariwal2021diffusion} as CADM for short.
Correspondingly, UADM means the unconditional diffusion architecture.
UADM-G and CADM-G additionally use noise-aware classifier guidance to perform conditional generation, separately.

Our training hyperparameters of noise-aware classifier, including batch size, total number of iterations, and decay rate, are kept the same with Dhariwal \etal \cite{dhariwal2021diffusion} for fair comparison.
We adopt the fixed linear variance schedule $\beta_1,...,\beta_T$ \cite{nichol2021improved,ho2020denoising} for prior noising process Eq.\eqref{eq:rw_forward} and choose $T$ as $1000$.
In this paper, we set $\eta$ to 0.2 to keep a slight disturbance during training stage of classifier in all experiments. 
The selection details of $\gamma$ can be seen in supplementary materials.
All the experiments are conducted on 16 NVIDIA 3090s.

\paragraph{\textbf{Evaluation Metrics.}}

We select FID (Fréchet Inception Distance) \cite{fid} as our default evaluation metric, which is the most widely used metric for generation evaluation.
FID \cite{fid} measures KL divergence of two Gaussian distributions, which is computed by the real reference samples and the generated samples, in the feature space of the Inception-V3. 
To capture more spatial relationships, sFID are prposed as a variant of FID, which is more sensitive to the consistent image distribution with high-level structures.

Besides, we apply several other metrics
for more comprehensive evaluations.
Inception Score (IS) was proposed \cite{noteoninceptionscore},
to measure the mutual information between input sample and the predicted class.
Improved Precision and Recall metrics are proposed \cite{precrecall} for further evaluation of generative models.
Precision is computed by estimating the proportion of generated samples that fall into real data manifold, measuring the sample fidelity.
By contraries, recall is computed by estimating the proportion of real samples which fall into generated data manifold, measuring the sample diversity.
Following Dhariwal.et al \cite{dhariwal2021diffusion}, we randomly generated \textbf{50k} images to compute all metrics above based on \textbf{10k} real images when compared with previous methods.
For consistent comparisons, we use the evaluation metrics for all methods based on the same codebase as Dhariwal \etal \cite{dhariwal2021diffusion}.

\subsection{Comparison with State-of-the-art Methods}

\begin{table}[t]
    \setlength\tabcolsep{2.5mm}
    \caption{Comparison results with state-of-the-art generative models on ImageNet1000 256$\times$256. Annotation '(25)' means the DDIM \cite{ddim} sampling method with 25 steps. Otherwise, it means DDPM sampling method with 250 steps.
    }\label{tab:sota}
    \begin{center}
    
    
    
    
    \begin{tabular}[t]{lccccc}
    
    \toprule
    Method            & FID $\downarrow$         & sFID  $\downarrow$ & IS $\uparrow$     & Prec  $\uparrow$    & Rec $\uparrow$ \\
    
    \toprule
    DCTransformer \cite{dctransformer} & 36.51 & 8.24 & - & 0.36 & \bf{0.67}  \\
    VQ-VAE-2 \cite{vqvae2}    & 31.11 & 17.38 & - & 0.36 & 0.57 \\
    IDDPM \cite{nichol2021improved}   & 12.26 & 5.42 & - & 0.70 & 0.62 \\
    SR3  \cite{sr3}  & 11.30 & - & - & - & -\\
    BigGAN-deep \cite{biggan}  & 6.95 & 7.36 & - & \bf{0.87} & 0.28\\
    \toprule
    UADM-G  \cite{dhariwal2021diffusion}(25)  & 14.21 & 8.53 & 83 & 0.7 & 0.46 \\
    \bf{UADM-G+ECT+EDS} (25) & 8.28 & 6.37 & 163.17 & 0.76 & 0.44 \\
    UADM-G  \cite{dhariwal2021diffusion}   & 12 & 10.4 & 95.41 & 0.76 & 0.44 \\
    \bf{UADM-G+ECT+EDS} & 6.78 & 6.56 & 168.78 & 0.81 & 0.45 \\
    \toprule
    CADM   \cite{dhariwal2021diffusion}     & 10.94 & 6.02 & 100.98 & 0.69 & 0.63 \\
    CADM-G (25)  \cite{dhariwal2021diffusion}  & 5.44 & 5.32 & 194.48 & 0.81 & 0.49 \\
    \bf{CADM-G+ECT+EDS} (25) & 4.67 & 5.12 & \textbf{235.24} & 0.82 & 0.47     \\
    CADM-G \cite{dhariwal2021diffusion} & 4.59 & 5.25 & 186.70 & 0.82 & 0.52 \\
    \bf{CADM-G+ECT+EDS} & \bf{4.09} & \bf{5.07} & 221.57 & 0.83 & 0.50 \\
    \toprule
    \end{tabular}
    
    \end{center}
    \vspace{-2pt}
\end{table}

In this section, we show the comparison results of our proposed schemes with other SOTA methods.
based on UADM and CADM  architectures \cite{dhariwal2021diffusion}.
All results of other previous methods are cited from Dhariwal \etal \cite{dhariwal2021diffusion}.
As shown in Table.\ref{tab:sota}, we achieve the best results in term of FID metric.
Compared with UADM-G in ImageNet 256$\times$256, our methods achieve relatively about 40\% improvement on FID metric (from 14.21 to 8.28, from 12.0 to 6.78) based on both DDPM and DDIM sampling iteration methods, with comparable or even better precision and recall.
When based on the CADM architecture, our proposed schemes still maintain a significant improvement margin on FID metric (from 5.44 to 4.67, from 4.59 to 4.09), and comparable precision and recall.

It is worth mentioning that our method based on UADM architecture outperformed BigGAN-deep in term of FID by 0.17 margin (6.78 vs 6.95), with no dependency on conditional architecture CADM, which has not been achieved in previous work \cite{dhariwal2021diffusion,ho2020denoising,nichol2021improved}.
\vspace{-6pt}

.
\subsection{Ablation Study}
\subsubsection{Effect of proposed schemes.}
\label{exp:overall_ab}
To further verify the contribution of each component of proposed schemes, we conduct ablation experiments on both UADM and CADM architectures, as shown in Table.\ref{tab:unconditional} and Table.\ref{tab:conditional}.
It can be concluded that EDS and ECT both improves the generation quality.
Combining with two schemes, the generation results can be further improved, with more semantic details achieved from improved direction (ECT) and scale (EDS) aspects of guidance gradient.

\subsubsection{Effect of entropy-driven sampling.}
\label{exp:eds_ab}
\begin{table}[!t]
    \caption{Effect of our proposed methods based on UADM-G under DDIM 25 steps and DDPM 250 steps on ImageNet1000 256$\times$256.}
    \label{tab:unconditional}
    \begin{center}
    \setlength{\tabcolsep}{1.8mm}{
    \begin{tabular}{cccccccc}
    \toprule
    DDIM & ECT & EDS & FID $\downarrow$   & sFID $\downarrow$ & IS $\uparrow$  & Precision $\uparrow$ & Recall $\uparrow$  \\
    \midrule
         &  &      & 12.0 & 10.4  & 95.41  & 0.76      & 0.44 \\
         & \cmark &   & 10.79 & 10.49 & 117.56  & 0.79      & 0.41 \\
         &   & \cmark  & 7.98 & 10.61 & \bf 178.73  & \bf 0.82      & 0.40 \\
         &  \cmark & \cmark & \bf 6.78 & \bf 6.56 & 168.78  & 0.81      & \bf 0.45 \\
    \hline
    \cmark     &  &      & 14.21 & 8.53  & 83  & 0.7      & \bf 0.46 \\
    \cmark     & \cmark &   & 12.21 & 8.14 & 100.95  & 0.74      & 0.44 \\
    \cmark     &  & \cmark   & 10.09 & 6.86 & 133.71  & 0.73      & 0.45 \\
    \cmark     &  \cmark & \cmark & \bf 8.28 & \bf 6.37 & \bf 163.17  & \bf 0.76      & 0.44 \\
    \bottomrule
    \end{tabular}
    }
    \end{center}
    \vspace{-2mm}
\end{table}
\begin{table}[!t]
    \caption{Effect of our proposed methods based on CADM-G under DDIM 25 steps and DDPM 250 steps on ImageNet1000 256$\times$256.}
    \label{tab:conditional}
    \begin{center}
    \setlength{\tabcolsep}{1.8mm}{
    \begin{tabular}{cccccccc}
    \toprule
    DDIM & ECT & EDS & FID $\downarrow$   & sFID $\downarrow$ & IS $\uparrow$  & Precision $\uparrow$ & Recall $\uparrow$  \\
    \midrule
        & &      & 4.59 & 5.25  & 186.7  & 0.82      & 0.52 \\
         & \cmark &   & 4.62 & 5.16 & 182.48  & 0.81      & \bf 0.53 \\
         & & \cmark  & \bf 4.01 & 5.15 & 217.25  & 0.82      & 0.52 \\
         & \cmark & \cmark & 4.09 & \bf 5.07 & \bf 221.57  & \bf 0.83      & 0.50 \\
    \hline
    
        \cmark &  &     & 5.46 & 5.32  & 194.48  & 0.81      & 0.48 \\
         \cmark &  \cmark &   & 5.34 & 5.3 & 196.8  & 0.81      & 0.49 \\
          \cmark &  & \cmark   & 4.82 & \bf 5.04 & 218.97  & 0.80      & \bf 0.50 \\
         \cmark & \cmark & \cmark &\bf 4.67 & 5.12 & \bf 235.24  & \bf 0.82      & 0.48 \\
    \bottomrule
    \end{tabular}
    }
    \end{center}
    \vspace{-4mm}
\end{table}
In this part, we design various intuitive scaling methods and compare them under optimal hyperparameters with EDS.
We select the scaling method with constant recovery factor for all time range \cite{dhariwal2021diffusion} as our baseline.

Intuitively, we can manually select vanishing time point through observing Fig.\ref{fig:vanishing_hist} and Fig.\ref{fig:coverge} i.e., 700, and finetune the constant rescaling factor to adjust the weak gradient for all generated samples, which we called Constant (range 0-700).
The constant scale can be further adjusted to time-aware form, such like $T-t$.
We call this method as Timestep-aware, which can dynamically adjust the scaling factor according to time step in sampling process.
In this way, the sample-aware vanishing characteristic is ignored and the scale cannot fit the various vanishing level.

To verify that entropy-aware scaling could better fit the initial time steps of vanishing guidance, we design another approach which is based on norm of gradient map.
Specifically, we empirically select a norm bound $M$ for gradient.
When the norm of gradient map is smaller than the threshold vanishing norm bound $M$, we regard that the gradient guidance is weak and need to be rescaled.
Thus, $s$ in Eq.\ref{eq:condsample} is rewritten as followed:
\begin{equation} \label{eq:entropy_clip}
\begin{split}
s = \left\{
\begin{array}{rcl}
1, & & {\Vert \nabla_{\bx_t} \log p_{\phi}(\by|\bx_t) \Vert_2 < M}\\
C,  & & \mathrm{Otherwise}\\
\end{array} \right.
\end{split}
\end{equation}
The rescaling factor $C$ is a large constant.
Compared with methods above, it can be verified experimentally that EDS not only fits the sample-aware initial time steps of weak guidance, but also provides reasonable rescaling factor for recovery, as shown in Table.\ref{tab:ablation_EDS}.


\begin{table}[t]
    \setlength\tabcolsep{3.5mm}
    \caption{Sample quality comparison with several intuitive sampling schemes. All results are evaluated under 5,000 generated samples for more efficient comparisons.}\label{tab:ablation_EDS}
    \begin{center}
    \begin{tabular}[t]{lcccc}
    \toprule
    Method            & FID          & IS       & Precision      & Recall \\
    \toprule
    Baseline  & 20.17        & 84.53          & 0.71          & 0.59       \\
    Constant (range 0-700)      & 19.84	       & 88.42          & 0.70	         & 0.61       \\
    Timestep-aware & 19.42    & 87.42          & 0.70	         & 0.63       \\
    Gradient Norm               & 19.08     & 93.14     & 0.71          & 0.63       \\
    Entropy-driven  & \bf 16.56    & \bf 133.26     & \bf 0.73          & \bf 0.66       \\
    \toprule
    \end{tabular}
    \end{center}
    \vspace{-1pt}
\end{table}
\begin{figure}[!t]
    \vspace{-3pt}
    \begin{center}
    \includegraphics[width=1.0\textwidth]{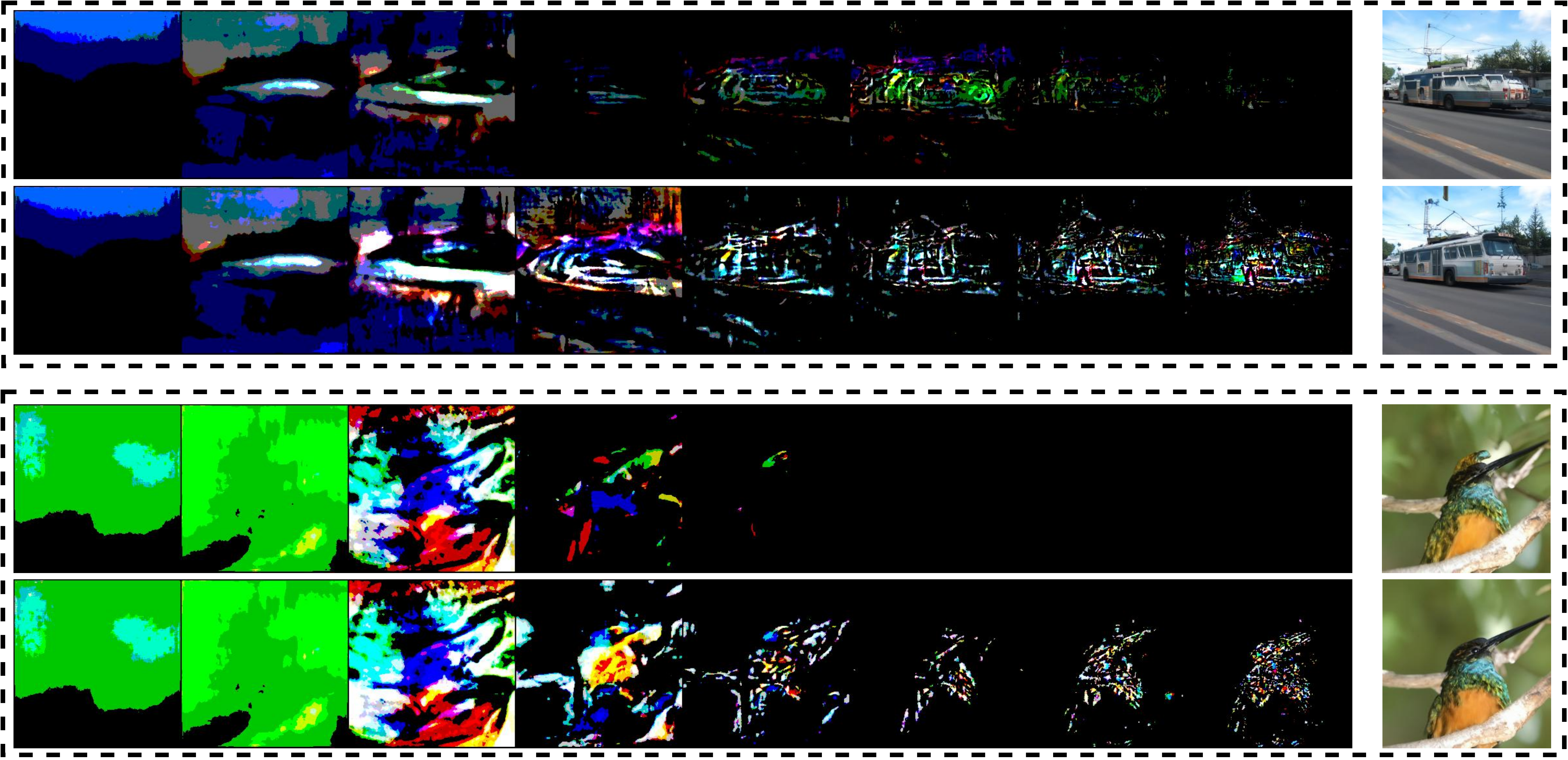}
    \caption{Gradient map comparison of UADM-G \cite{dhariwal2021diffusion} (the first line of each image box) and our UADM-G + EDS (the second line of each image box) on ImageNet1000 256$\times$256 using DDIM 25 iterations. For better visualization, we pick 8 maps at the same interval from 25 consecutive maps and magnify all pixels of a image by a factor of 50.
    }\label{fig:grad_vis}
    \end{center}
    \vspace{-8mm}
\end{figure}

\begin{figure}[!t]
     \centering
     \begin{subfigure}[b]{0.32\textwidth}
         \centering
         \includegraphics[width=\textwidth]{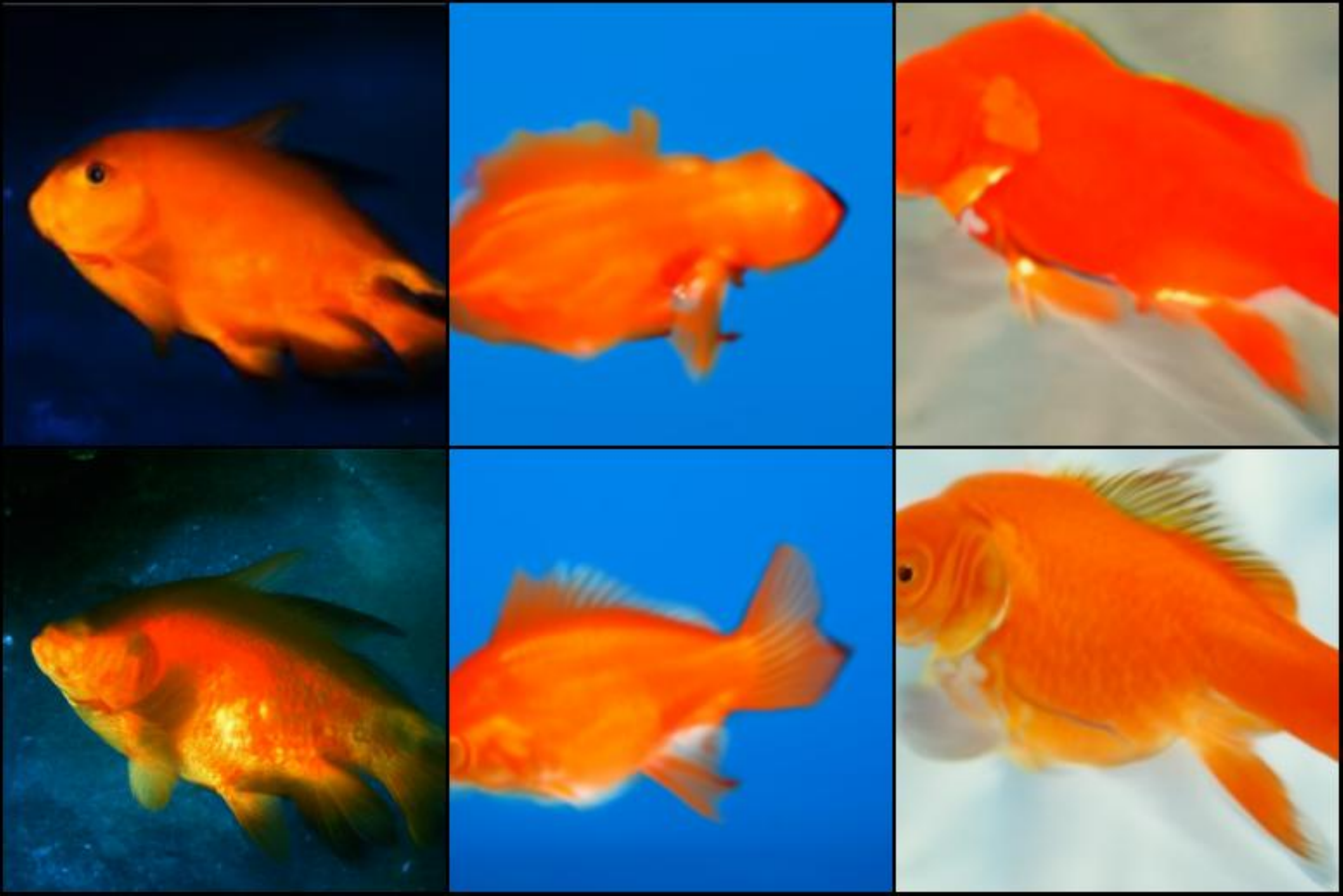}
         \caption{Class 1:goldfish}
         \label{fig:compare_0}
     \end{subfigure}
     \hfill
     \begin{subfigure}[b]{0.32\textwidth}
         \centering
         \includegraphics[width=\textwidth]{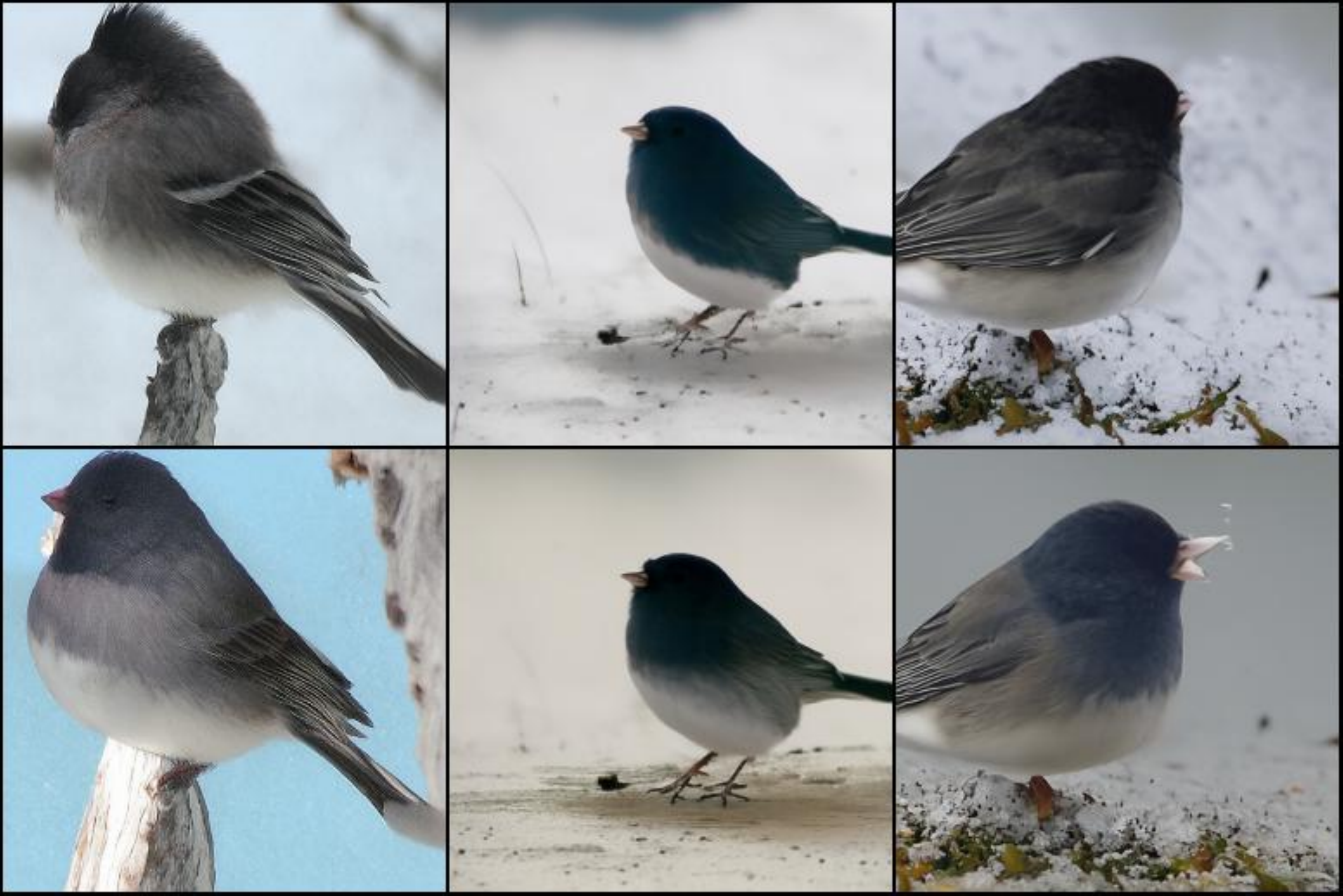}
         \caption{Class 13:junco}
         \label{fig:compare_1}
     \end{subfigure}
     \hfill
     \begin{subfigure}[b]{0.32\textwidth}
         \centering
         \includegraphics[width=\textwidth]{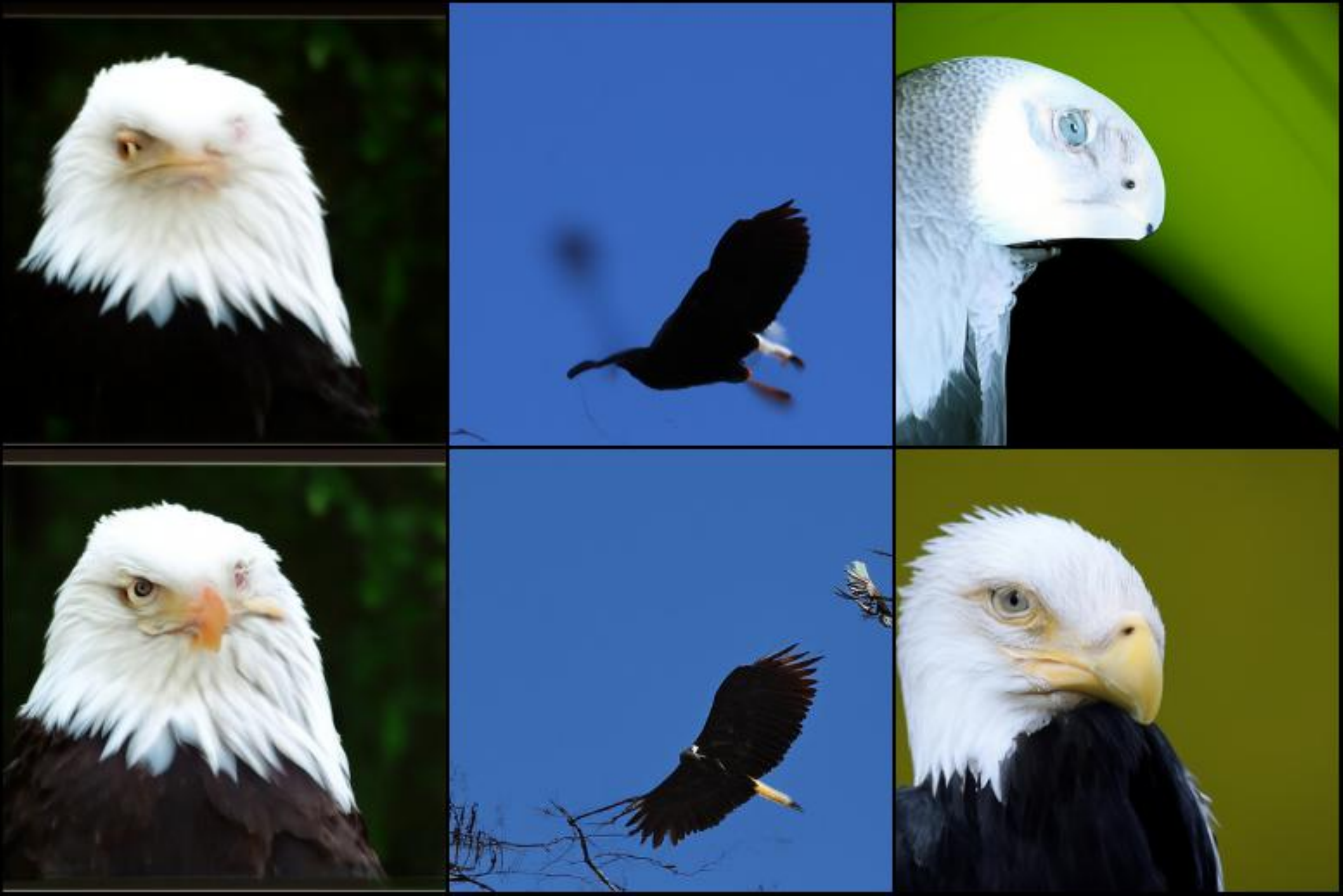}
         \caption{Class 22:eagle}
         \label{fig:compare_2}
     \end{subfigure}
     \hfill
     \begin{subfigure}[b]{0.32\textwidth}
         \centering
         \includegraphics[width=\textwidth]{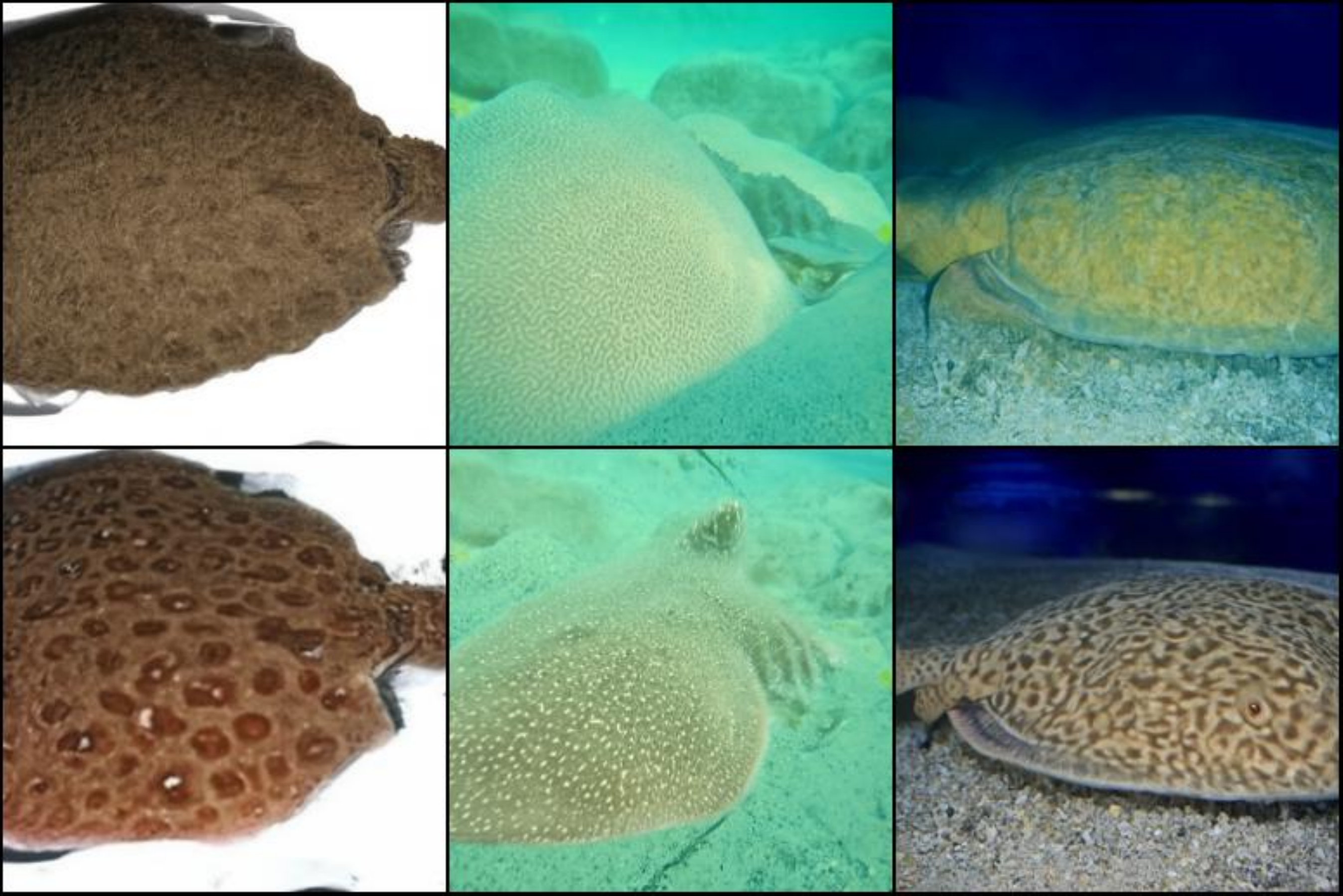}
         \caption{Class 5:numbfish}
         \label{fig:compare_3}
     \end{subfigure}
     \hfill
     \begin{subfigure}[b]{0.32\textwidth}
         \centering
         \includegraphics[width=\textwidth]{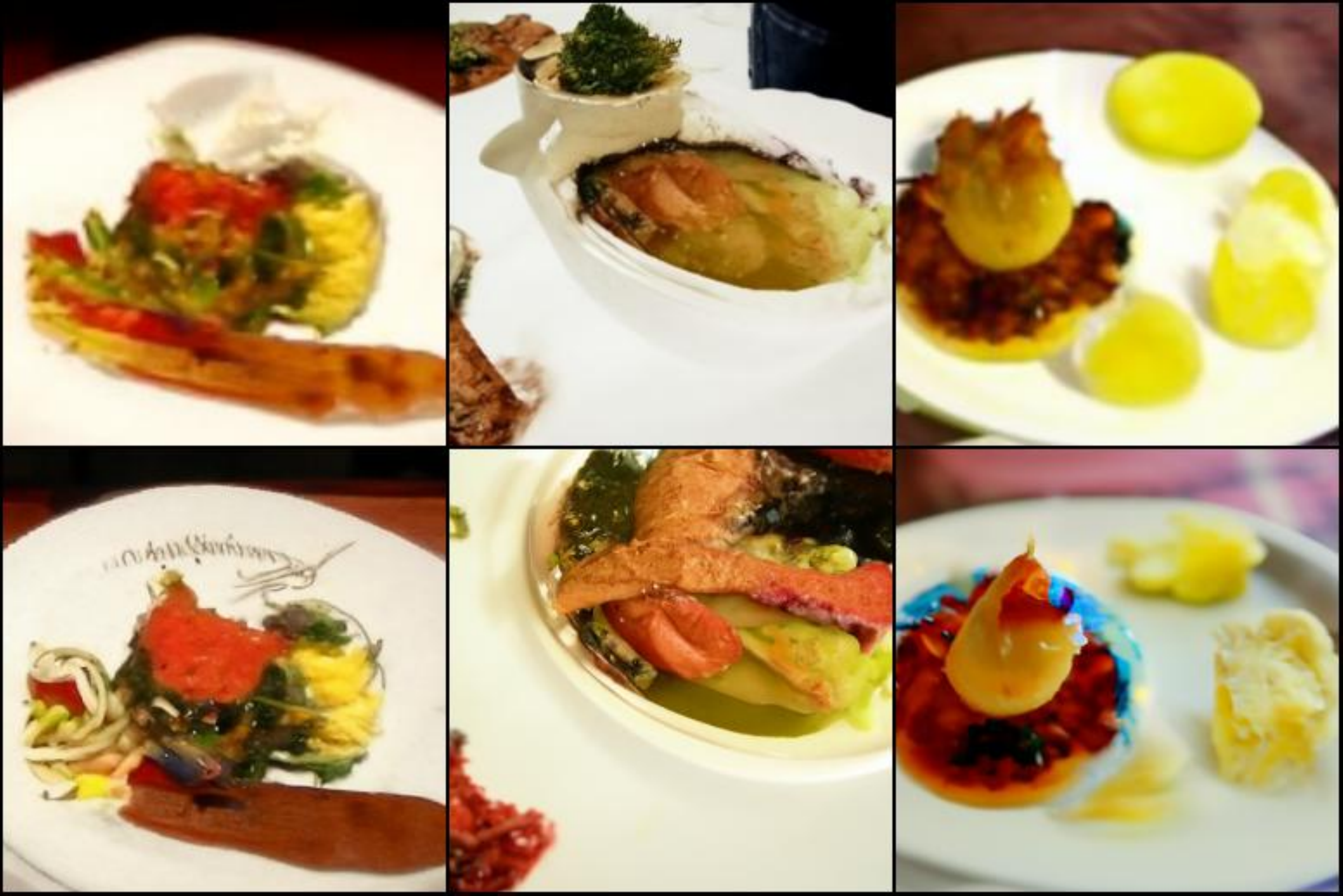}
         \caption{Class 923:plate}
     \end{subfigure}
     \hfill
     \begin{subfigure}[b]{0.32\textwidth}
         \centering
         \includegraphics[width=\textwidth]{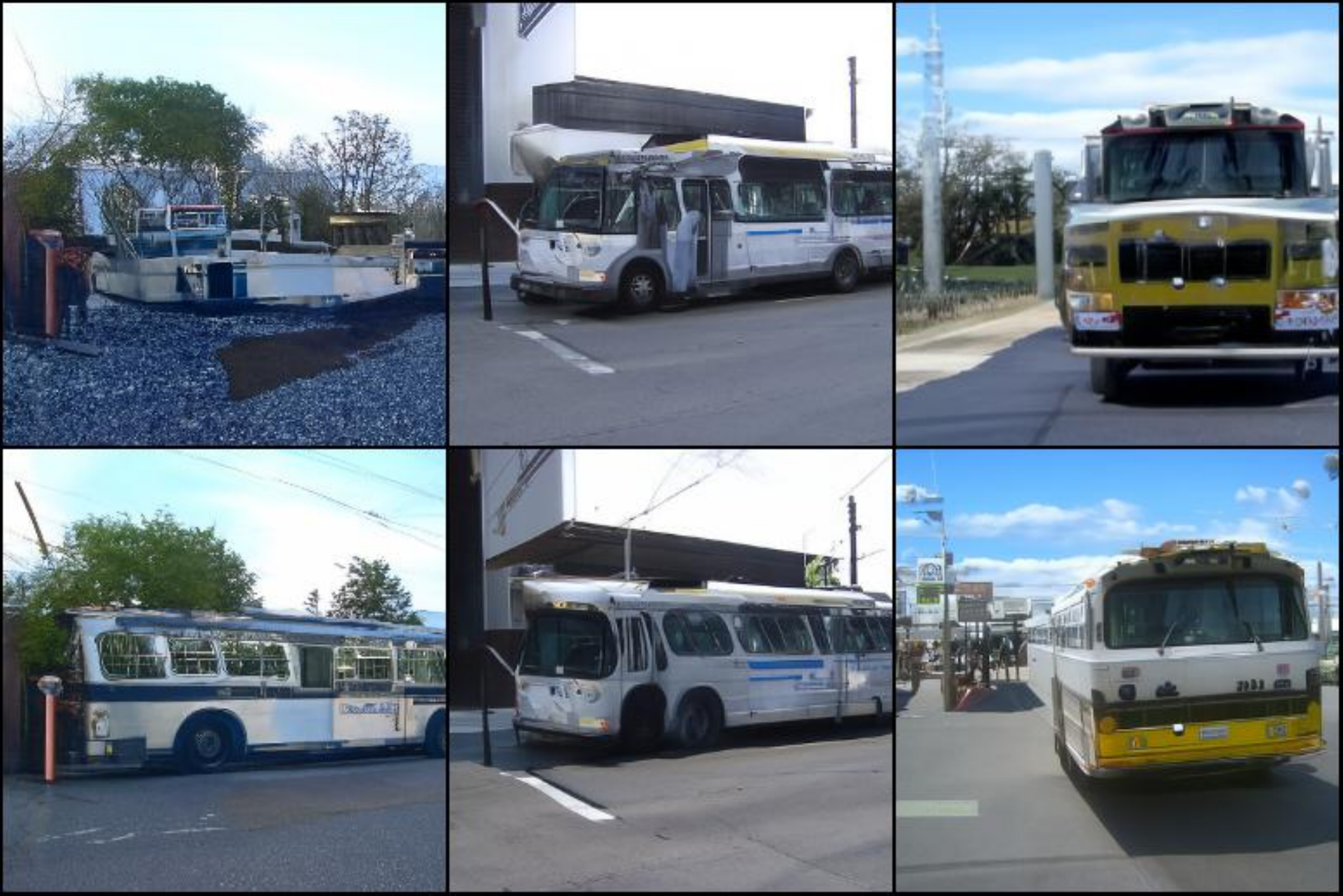}
         \caption{Class 874:trolleybus}
         \label{fig:compare_5}
     \end{subfigure}
    \begin{subfigure}[b]{0.32\textwidth}
         \centering
         \includegraphics[width=\textwidth]{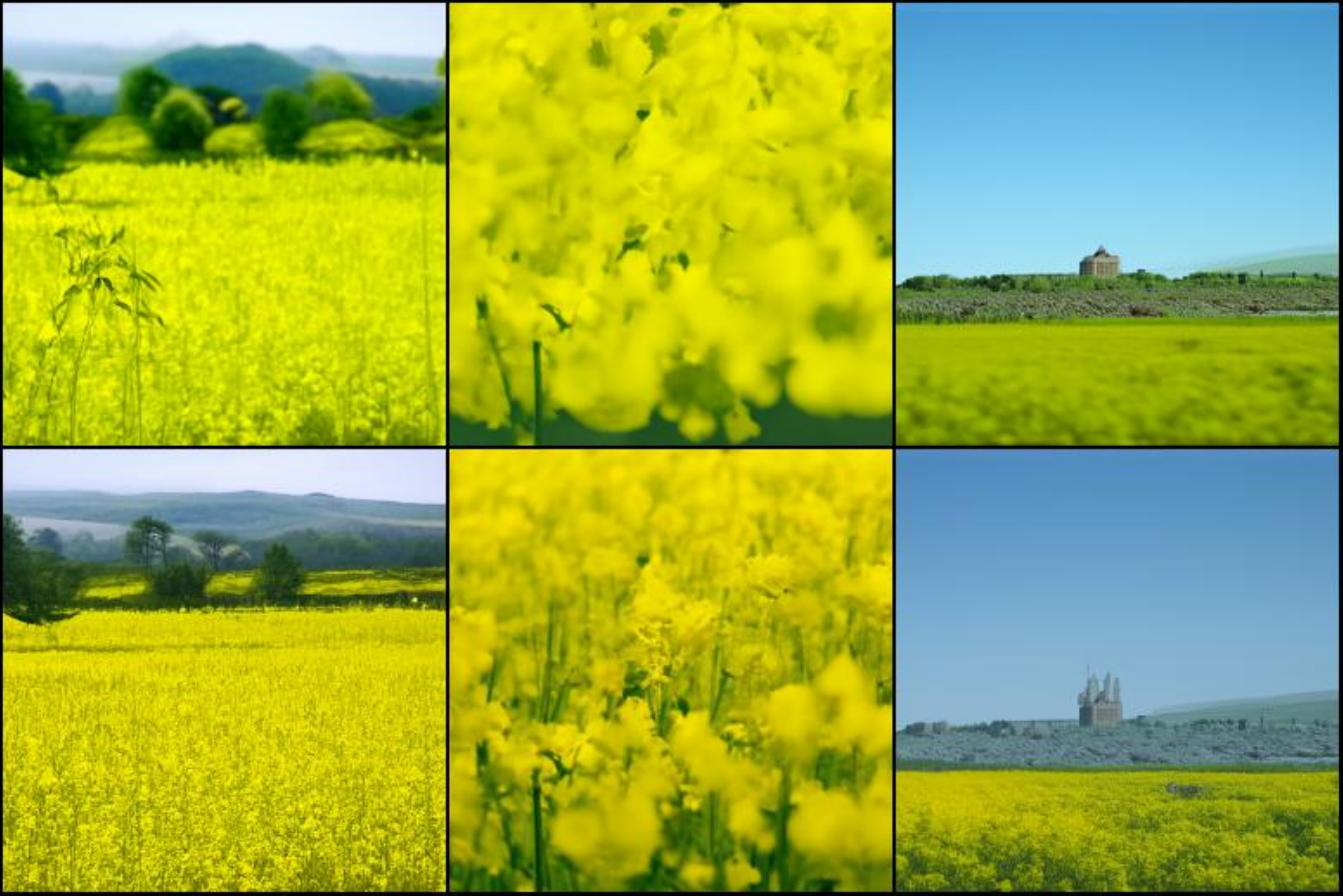}
         \caption{Class 984:rapeseed}
         \label{fig:compare_6}
     \end{subfigure}
     \hfill
     \begin{subfigure}[b]{0.32\textwidth}
         \centering
         \includegraphics[width=\textwidth]{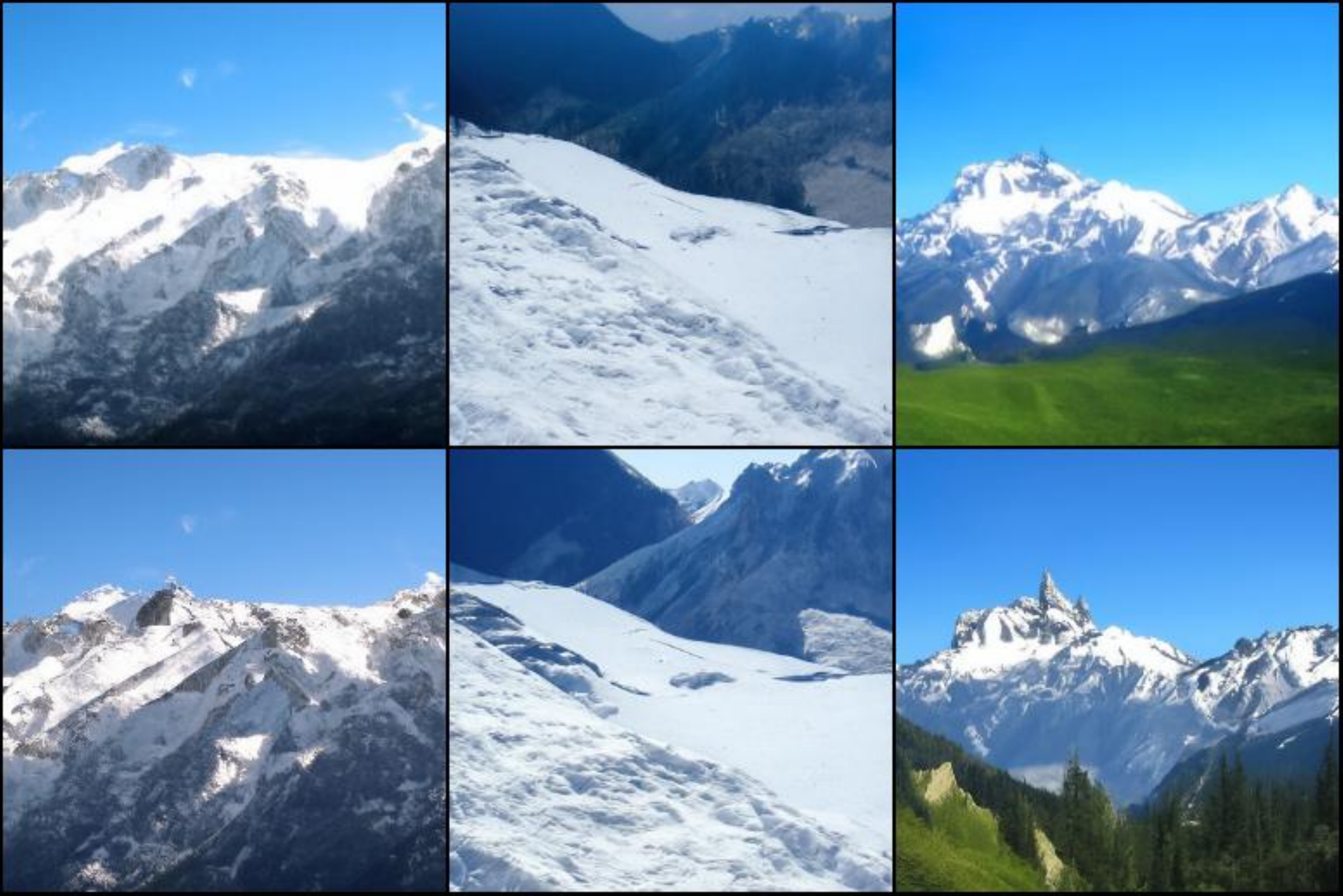}
         \caption{Class 970:alp}
         \label{fig:compare_7}
     \end{subfigure}
     \hfill
     \begin{subfigure}[b]{0.32\textwidth}
         \centering
         \includegraphics[width=\textwidth]{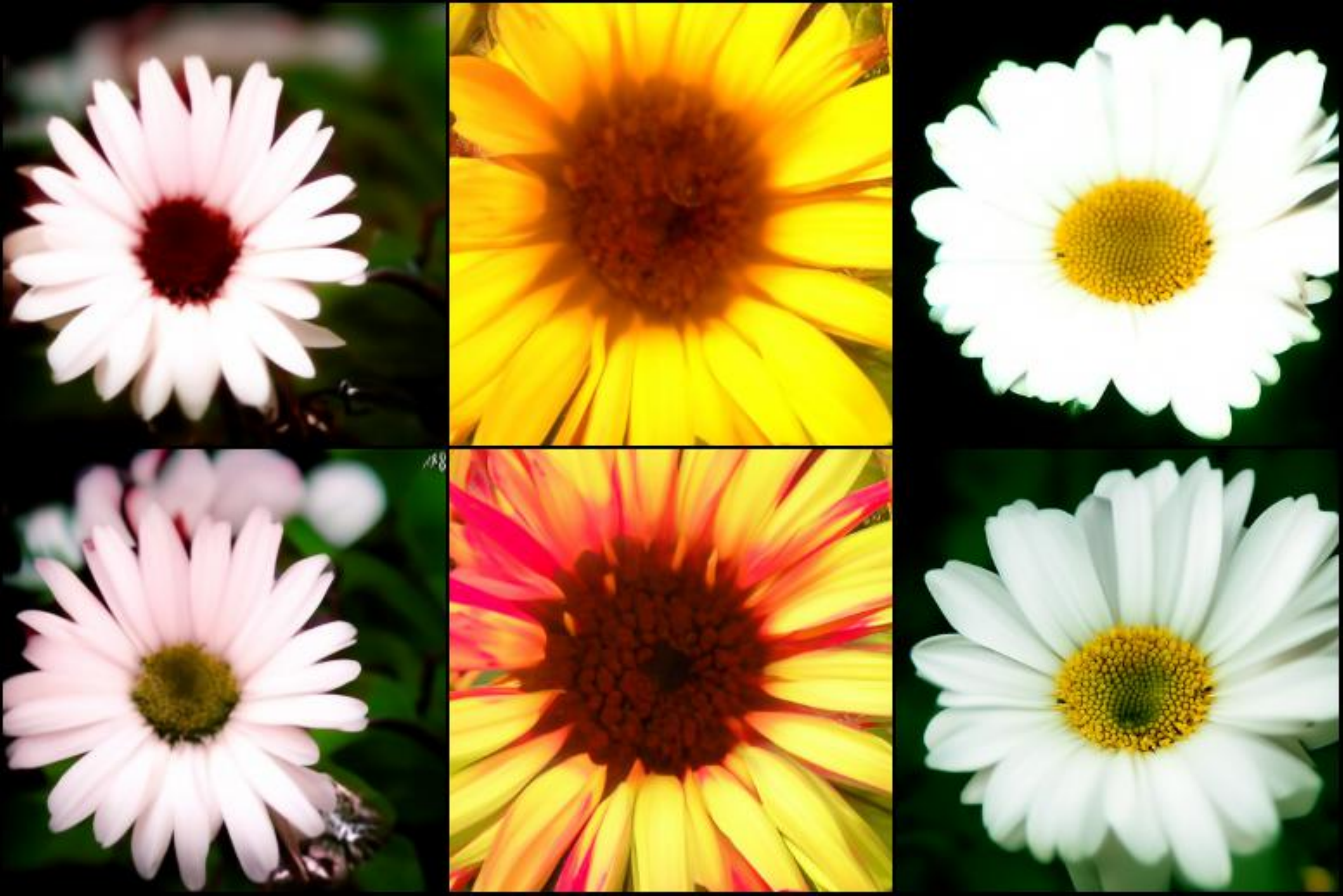}
         \caption{Class 985:daisy}
         \label{fig:compare_8}
     \end{subfigure}
    \caption{
    Sample quality comparison of UADM-G \cite{dhariwal2021diffusion} (the first line of each image box) and our UADM-G + ECT + EDS (the second line of each image box) on ImageNet1000 256$\times$256 using DDIM 25 iterations. 
    }
    \label{fig:comparation_results}
    \vskip -0.2in
\end{figure}
\subsection{Qualitative Results}
\subsubsection{Gradient Map Visualization.}
In this part, we collect several gradient maps $\bg$ derived from classifier in the previous sampling process and our EDS process, with an equal time interval.
From Fig.\ref{fig:grad_vis}, it can be observed that the classifier provides high-level semantic guidance at the beginning.
Gradually, the classifier will provide the condition-aware texture guidance.
Compared with EDS, which can maintain guidance of semantic details throughout the denoising process for refined generation results, sampling scheme based on the fixed scaling factor lost a lot of condition-aware details (first line) or introduce unnatural details (third line) at the later sampling stage.


\subsubsection{Generation Results Visualization.}
In this part, we visualize various generated images conditioned on different classes.
Based on UADM architecture, we compared our final generated results with that of previous SOTA method \cite{dhariwal2021diffusion}.
Considering the stochastic noise involved in ddpm sampling process, we adopted DDIM, the deterministic generation process, to generate samples with shared initial noise.
From Fig.(\ref{fig:comparation_results}), previous SOTA method can not generate condition-aware semantic details such like beaks of birds Fig.(\ref{fig:compare_1}) or petal texture Fig.(\ref{fig:compare_8}), while our method is able to generate more refined textures, with similar high-level structure.
\section{Conclusion}
In this paper, we proposed an entropy-aware scaling technology for the guiding classifier in sampling process.
From the perspective of training, we propose an entropy-aware optimization loss to constrain the predicted distribution, alleviating the overconfident prediction for noisy sample in the sampling process.
Experiments demonstrate that our proposed methods can recover more textures in generated samples,
achieving state-of-the-art generation results.
\bibliographystyle{splncs04}
\bibliography{eccv2022}
\end{document}